\documentclass[runningheads]{llncs}
\usepackage{graphicx}
\usepackage{amsmath}
\DeclareMathOperator*{\argmax}{arg\,max}

\graphicspath{{./images/}}
\begin{document}
\title{NBcoded: network attack classifiers based on Encoder and Naive Bayes model for resource limited devices}

%
%\titlerunning{Abbreviated paper title}https://www.overleaf.com/project/60a248b59374c464c9c3f9cf
\author{Lander Segurola-Gil\inst{1} \and
Francesco Zola\inst{1,2} \and
Xabier Echeberria-Barrio\inst{1} \and Raul Orduna-Urrutia\inst{1}}
\authorrunning{L. Segurola et al.}
\institute{Vicomtech Foundation, Basque Research and Technology Alliance (BRTA) \email{\{lsegurola, fzola, xetxeberria, rorduna\}@vicomtech.org} \and
Institute of Smart Cities, Public University of Navarre, 31006 Pamplona, Spain
}
\maketitle              % typeset the header of the contribution
\begin{abstract}
In the recent years, cybersecurity has gained high relevance, converting the detection of attacks or intrusions into a key task. In fact, a small breach in a system, application, or network, can cause huge damage for the companies. However, when this attack detection encounters the Artificial Intelligence paradigm, it can be addressed using high-quality classifiers which often need high resource demands in terms of computation or memory usage. This situation has a high impact when the attack classifiers need to be used with limited resourced devices or without overloading the performance of the devices, as it happens for example in IoT devices, or in industrial systems. 
For overcoming this issue, NBcoded, a novel light attack classification tool is proposed in this work. NBcoded works in a pipeline combining the removal of noisy data properties of the encoders with the low resources and timing consuming obtained by the Naive Bayes classifier. This work compares three different NBcoded implementations based on three different Naive Bayes likelihood distribution assumptions (Gaussian, Complement and Bernoulli). Then, the best NBcoded is compared with state of the art classifiers like Multilayer Perceptron and Random Forest. Our implementation shows to be the best model reducing the impact of training time and disk usage, even if it is outperformed by the other two in terms of Accuracy and F1-score ($\sim$ $2$\%). 

%Cybersecurity has become a field of high relevance due to the number of devices connected to the internet and the data these generate. A critical challenge when talking about cybersecurity is the detection of attacks or intrusions. This problem is often tackled with high resource demanding detectors. For overcoming this issue, NBcoded, a novel light attack classification tool is proposed in this work. NBcoded works in a pipeline combining an encoder with Naive Bayes classifiers, analyzing network data by firstly comprising it and getting relevant patterns of it, and then classifying it into a normal or attack category. Comparing the presented approach to other state-of-the-art methods, really competitive results are obtained in terms of performance, outperforming those in terms of needed resources.
\end{abstract}
\section{Introduction}
 In the midst of the era of the Internet of Things (IoT), cybersecurity is becoming a field with strong importance in the daily life. In fact, cyber-attacks are evolving incredibly fast, making them more sophisticated \cite{sallinen2021weaponized} as time goes by. Not only that, the domain where these can be deployed is increasing at a significant velocity due to the fast growth of devices connecting to the internet \cite{lombardi2021internet}. According to \cite{lombardi2021internet}, there are about $20$ billion devices connected to the global net. These are not only composed of wearables; appliances from medical devices to automotive control units are part of the huge net thanks to the Industry $4.0$ paradigm, the Smart Factory and the Industrial IoT (IIoT). In fact, many industries and critical infrastructures have now many devices connected to the internet \cite{boyes2018industrial}. Therefore, the consequences of a cyber-attack are not only limited to digital leak or losses, hijacking, Denial of Service (DoS) or ransom, but a much larger scope, affecting in economical, reputational, psychological and societal terms too \cite{Agrafiotis2018ATO}.

Due to the need of confronting these cyber-threats, cybersecurity has evolved too. Many approaches have been proposed in order to avoid or alert when a cyber-attack is going on. Initially, static rule based solutions could work, but as it is mentioned, with the evolution of the cyber-threads, Artificial Intelligence (AI) based solutions \cite{wirkuttis2017artificial} burst strongly. When trying to detect or predict anomalous situations, Machine Learning (ML) techniques, a branch of AI, can provide specially interesting approaches \cite{verbraeken2020survey}. These algorithms allow network suitable tools that may learn specific patterns.
%which makes the choice of these learning techniques particularly appropriate for this scenario \cite{buczak2015survey}. 

When talking about limited resourced environments, these may not be capable of assuming the requirements that many of these AI algorithms need, in terms of resources and time consumption \cite{krishnan2019artificial}. For example, in some IoT devices and Industrial systems, several problems might end up being faced when trying to implement one of these AI based solutions, with regard to the need of huge computational power, lack of enough RAM, or even lack of disk space for saving models. This scenario is not an appropriate one to deploy a high demanding Deep Learning (a branch of ML) model, and there will be the need of a trade off resources for accuracy. In fact, this will become a challenge for the AI systems \cite{krishnan2019artificial}.

This problem should be tackled by the creation of light learning models, in spite of making sacrifices in terms of performance, while trying to maintain it as much as possible, in exchange of a low resource usage. For this reason, in this work, we present a novel light and fast attack classification system, with the name of NBcoded. The main idea is to exploit the lightness and fastness of the Naive Bayes (NB) classifiers, while exploiting the autoencoders abilities to deconstruct and construct the initial data, in order to reduce the noisy data and get more clear patterns of it. This, enhances the classification task for the NB classifiers, preserving its lightness. NBcoded works by combining  autoencoder and NB technologies, aiming to classify network traffic into attack/unexpected or normal traffic. 

The main goals of this work are:
\begin{itemize}
    \item Compare Naive Bayes classifiers assuming different distributions, comprising Gaussian, Complement and Bernoulli, in a network traffic classification task.
    \item Improve the Naive Bayes classification quality by adding an extra-layer based on Autoencoder information.
    \item Compare the best results of the previous experiments with state of the art techniques, in terms of quality, training time and disk usage.
\end{itemize}

The rest of the paper is divided as follows: in Section \ref{Background}, general concepts as well as state of the art are introduced. Section \ref{Methodology} explains the proposed approach for attack classification, detailing each member composing the NBcoded. In section \ref{ExpFram}, the used dataset as well as the selected evaluation metrics are introduced. Section \ref{ExpSt} presents the performed experiments and section \ref{Results} illustrates the results obtained in those. Finally, section \ref{Conc} presents the conclusions of this work as well as the guidelines for future works are drawn.

%Related to Deep Learning (DL) applied to cybersecurity, much information can be found in the review \cite{dixit2021deep}. 

\section{Background}\label{Background}
In this section, several general concepts are introduced, in particular in Section \ref{NBc} and Section \ref{AE}, the main concepts of the Naive Bayes classifier and Autoencoders (AE) are introduced respectively, and in Section \ref{RW} the state of the art is presented.
\subsection{The Naive Bayes Classifier} \label{NBc}
The Naive Bayes (NB) classifier \cite{murphy2006naive} is a probabilistic classifier based on the Bayes theorem \cite{bayes1763lii}, together with the next assumption; Let $(X_1,...,X_n)$ be a vector of variables and let $C$ be a dependant variable on $X_i$ for $i=1,..,n$. Then the NB classifiers naively assumes the conditional independence for all $X_i$ with $i=1,...,n$ related to $C$. In other words,
\begin{equation}\label{eq:NB}
    P(C|X_1,...,X_n)=\frac{P(X_1,...,X_n|C)P(C)}{P(X_1,...,X_n)}=\frac{P(C)\Pi_{i=1}^{n} P(X_i|C)}{P(X_1,...,X_n)},
\end{equation}

where $P(C|X_1,...,X_n)$ is called the posterior probability, $P(C)$ is the prior probability, and $P(X_1,...,X_n)$ is the evidence. This last part, is a constant that can be calculated by the provided evidence or data, so
$$
P(C|X_1,...,X_n)\propto P(C)\Pi_{i=1}^{n} P(X_i|C).
$$
Then, basically, given an observation $(x_1,...,x_n)$ and a NB classifier $f$, for a set of classes $C$, $f$ will make the prediction by using the Maximum a posteriori (MAP) estimation, by
$$
f(x_1,...,x_n)=\argmax_{c}P(C=c)\Pi_{i=1}^{n} P(X_i=x_i|C=c).
$$

To completely define the NB classifier, it is needed to determine $P(X_i|C)$ distributions. For this, many distribution can be assumed, such as the Gaussian distribution, Bernoulli distribution \cite{mccallum1998comparison} or Multinomial distribution \cite{rennie2003tackling}. From this procedure the NB learning algorithm is created, and from this point and for the rest of the paper, depending on the chosen distribution, the model will be called the Gaussian NB (GNB), Bernoulli NB (BNB) or Complement NB (CNB) \cite{rennie2003tackling}, where the last one is a derivation of the Multinomial distribution for imbalanced datasets.

\subsection{Autoencoders}\label{AE}

\begin{figure}[]
    \centering
    \includegraphics[scale=0.7]{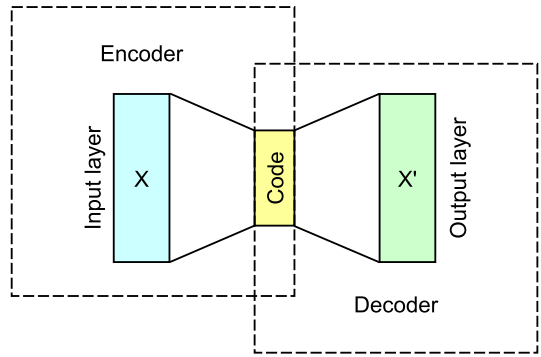}
    \caption{Autoencoder basic structure}
    \label{fig:auto}
\end{figure}

The idea of autoencoder was firstly introduced in \cite{bourlard1988auto}. These learning algorithms are a kind of Neural Networks (NN), created with the aim of encoding or representing data, commonly for dimension reduction. The architecture of an autoencoder may differ depending on the amount of hidden layers, but basically they all have the same structure. They are compounded by an input layer, gradually smaller hidden layers (where the last one is called the bottleneck), gradually bigger hidden layers and the output layer, which has the same size as the input layer (Figure \ref{fig:auto}).

 In the training phase, the model learns to compress the data and to decompress it. Thus, the \textit{encoder} component (Fig.\ref{fig:auto}) learns to condense the data, maintaining the relevant information, so the \textit{decoder} component (Fig.\ref{fig:auto}) can recover the original data. Once the model is trained, the encoder component reduces the dimension of the data, compressing the relevant information and removing the noise of the data, i.e, the irrelevant or duplicated information. 

\subsection{Related Work} \label{RW}
The AI, in particular, the ML branch, has been widely studied in the recent years. For example, in \cite{burnaev2016one,zhou2006extraction} a One-class Supported Vector Machine (SVM) for malware detection and for brain tumor detection respectively is used, in \cite{khoo2021solving} some problems related to partial differential equations are solved using Artificial Neural Networks (ANN), in \cite{sundararajan2021sleep} sleep classification is performed by the application of Random Forest (RF). Related to cybersecurity, in \cite{shaukat2020performance}, a review where different ML techniques composed by the ones mentioned above and other such as the Naive Bayes (NB), are compared against different tasks like intrusion, malware and spam detection. Last years Deep Learning (DL) architectures have provided a great leap forward in ML field. Many DL techniques have been studied in \cite{dixit2021deep} in attack detection tasks. 

In particular, the NB classifier is widely studied on many fields. For example, in medicine for the detection of cerebral infarction \cite{rukmawan2021cerebral}, or for coronary heart disease, breast cancer and diabetes \cite{jackins2021ai}. Research in many other fields can be found too; in \cite{wang2021using}, for predicting water floods, or in \cite{menon2020performance}, for earthquake predictions. Another interesting study is presented in \cite{li2020automatic}, where they propose an automatic bridge crack recognition tool based on CNN and NB.

Regarding cybersecurity, in \cite{belouch2018performance} NB along with other ML algorithms are tested against the UNSW-NB15 dataset. In \cite{david2020novel}, they propose an Intrusion Detection System (IDS), using the NB and SVM as reference models with which to test against, using the NSLKDDCup1999 \cite{verma2020review} dataset.

For Autoencoders, in \cite{mienye2020improved}, this technology is used for prediction of hearth disease, in \cite{lyons2014predicting} for the prediction of C$\alpha$ angles and dihedrals from protein sequences or in \cite{saad2018autoencoder} for explosion and earthquake classification.

In the cybersecurity field, autoencoders are used in \cite{ahmed2019mitigating} for mitigating covert cyber attacks in smart grids, or fooling IDS like in \cite{chen2020fooling}. In particular, for attack detection, in \cite{catak2019distributed} autoencoders are used for Distributed Denial of Service (DDoS) detection or in \cite{al2018deep}, where they combine it with a SVM for a binary and five different attack class classification.

\section{Proposal}\label{Methodology}

In this section, the NBcoded pipeline is presented. In particular, in Section \ref{Proposal}, the architecture of the presented classifier is presented and in Section \ref{netbe} the learning process is clarified.

\subsection{NBcoded Architecture}\label{Proposal}
The family of NB algorithms is a technology that has been widely studied  in the state-of-the-art as for example in \cite{verma2020review,david2020novel}. They bring light and fast classification models, but they often do not get high performance when compared to some other state-of-the-art methods. In some cases, this might be due to the incapacity of these naive methods to reach deeper patterns, despite having the advantage of being very light and needing a minimum request of resources for the processing, particularly when they face other training models. The main idea of this work is to present NBcoded, a light, and fast attack classification tool, which combines the lightness of the NB models with an enhanced performance given by the autoencoders, preserving the need for a limited quantity of resources, but bringing a competitive performance.
 
\begin{figure}[]
    \centering
    \includegraphics[scale=0.5]{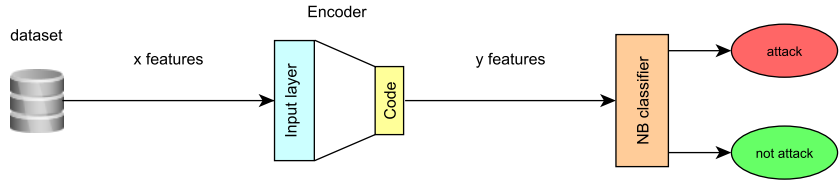}
    \caption{NBcoded arquitecture}
    \label{fig:NBenc}
\end{figure}

As shown in Figure \ref{fig:NBenc}, NBcoded consists of two parts; the data encoder part and the NB based attack classification part. The introduced model works by analyzing data flows in two steps: firstly,  the flows given data is  encoded into a less dimensional space. Then, the codification is given to the NB algorithms to be classified.  
 The idea of applying an encoder is making an improvement in classification tasks for the NB classifiers, making data less noisy and projecting data to get fewer features. 
 
 %and consequently, less dependency between them. In fact, the pipeline makes the NB methods getting much better results as it is presented in Section \ref{Results}, without needing lots of resources for it, making the tool appropriate for several limited environments.
 
 %\subsection{Component Architecture}\label{pipe}
 %As explained above, the encoder is the responsible of carrying out the first phase. The encoder is composed of the descending layers of the autoencoder. Thus, for the generation of one, it is needed to train an autoencoder first. Data will be given to the autoencoder, so it learns to encode it in the first part of its structure, for then decode it in the second one. The training of the autoencoder will provide a trained encoder which will  have potentially learnt to reduce noise in the data while compressing it in a less dimensional space. In this point, the encoder is subtracted and it is placed in the first part of the pipeline. As it is mentioned above, when deploying this part in a real scenario, the aim of the encoder will be to clean and simplify the data so the second part can perform in an enhanced fashion. 

For the first part, the creation of the encoder, an autoencoder is created first. For this task, a multilayer autoencoder is choosen, that is, an autoencoder with more than one hidden layer. More specifically, the chosen architecture is constructed as it follows; due to the selected features, the input layer consists of 9 features. Then, it is followed by a 8 neuron hidden layer, to end descending into a 6 neuron layer. Symmetrically, it ascends until the output layer is reached. The chosen optimizer has been the \textit{Adam} optimizer, with the mean absolute error loss and an l2 regularizer with a factor of $0.001$. With regard to the activation function, the \textit{tanh} has been the one choosen for every layer. From this structure, the encoder layers of an autoencoder are taken, that is, the 9 neuron, 8 neurons and the 6 neurons layers. The task of this part will be to encode data of dimension 9, to a dimension of 6, depicting the most representative patterns in the new space, cleaning it from noisy points.
 
 For the second part, the output of the first part will be taken as the input of the classifier with the task of classifying attacks. For this, the NB classifier is proposed. 
 
 The composition of these two technologies results in a light and fast classifier. In fact, the selected autoencoder structure is composed by a little amount of layers/neurons, what makes the encoder even lighter. The Naive Bayes classifiers are well known by their lightness and fastness too.
 
 \subsection{NBcoded learning}\label{netbe}
 In this architecture, a 9 neurons input layered autoencoder will receive a 9 featured dataset for the training phase. The training set will be split into two same sized parts. One of both will be used in the mentioned training phase of the autoencoder. Once the training is done, the trained encoder will be subtracted from it. At this point, the encoder will receive the other part with the task of encoding it. 
 
 For the second part, the NB classifier will receive the output of the encoder, comprising 6 featured points, so it can be trained. In this case, an assumption between Gaussian, Complement and Bernoulli distributions will be done. Firstly, the prior is calculated from the data points, getting the probability of a given random point belonging to one class or the other. Then, the prior will be updated as it follows; In a first step, the assumed likelihood distributions parameters will be fitted by the provided encoded data, and then the prior will be updated following the Equation \ref{eq:NB}, resulting in the a posteriori distribution. Once the training is done, the new distribution will remain static with no updates.

 Following with the real scenario case, once the proposed tool is ready for classification, it will receive 9 featured data points as input, that will be projected into a 6 dimensional space. Then the encoded data points will be classified into a normal class (0) or or attack/unexpected class (1), giving a binary classification of each point as the output.
 
 %the chosen model will receive the compressed data given by the encoder for a proper analysis. The output of the classifier will compress two options; it will classify the data flows into normal class (0) or attack class (1). In this work, three priors are tested, comprising the Gaussian, Complement, and Bernoulli priors. Depending on the used prior, the tool will be referred to as Gaussian NBcoded, Complement NBcoded, or Bernoulli NBcoded.

 %For this, three NB classifiers are tested, the Gaussian NB, Complement NB and Bernoulli NB. The task given will be to classify into a normal class (0) or attack class (1). In the three cases, priors are not given, so they are adjusted from the data. 
 
\section{Experimental Framework}\label{ExpFram}
In this section, an explanation of the used dataset is given, as well as the evaluation metrics selected. In Section \ref{DO}, the used dataset is explained, in Section \ref{DP}, the selected features together with the carried preprocess are presented, in Section \ref{EM} the evaluation metrics are explained, and finally, in Section \ref{MLap}, the chosen ML parameters, as well as used computer characteristics, are shown.

\subsection{Dataset Overview}\label{DO}
The experiments drawn in this research are implemented using the UNSW-NB15 dataset\footnote{https://www.unsw.adfa.edu.au/unsw-canberra-cyber/cybersecurity/ADFA-NB15-Datasets/}, a complete network flows dataset created with the aim of covering some gaps found in other benchmark datasets \cite{moustafa2015unsw}. This dataset is widely used in cybersecurity, for example for creating a deep learning binomial classifier based on neural networks for implementing an intrusion detection system \cite{al2017experimental}; for presenting a framework that combines linear and learning algorithms to create a Hybrid Anomaly Detection Model (HADM) \cite{monshizadeh2019performance} or to introduce a Deep Feature Embedding Learning (DFEL) framework for anomaly prediction which combines neural networks for feature embedding and making predictions \cite{zhou2018deep}. It is also used for extracting temporal graph information that is then used for classifying network behaviours, as introduced in \cite{zola2021temporal}.

The UNSW-NB15 dataset contains real normal and synthetic abnormal network traffic, generated in the University of New South Wales (UNSW) cybersecurity lab. In particular, the abnormal traffic is generated by deploying $9$ different attack families, which are Fuzzers,  Analysis, Backdoors, Denial-of-Service (DoS), Exploits, Generic, Reconnaissance, Shellcode and Worms.
The whole dataset is generated and divided into two separated captures, the first one of $16$ hours, and the second of $15$ hours. Then, the whole data traffic is preprocessed using tools like Argus, and Bro-IDSArgus \footnote{https://qosient.com/argus/index.shtml} and Bro-IDS\footnote{https://zeek.org/} for aggregating similar flows into unique records defined by $47$ features and two labels fields, one for indicating normal/attack flows (binary label), and the other one for specifying the attack family among the $9$ available. This aggregation reduces the two capture days up to $12$ hours and $12$ hours and a half, respectively, with $2,540,044$ number of samples of which $2,218,761$ labelled as normal connections and $321,283$ labelled as attack connections.

\subsection{Dataset preprocess and feature selection}\label{DP}
Even if the selected dataset was made with the aim of covering limitations, several studies, as \cite{zoghi2021unsw} have shown that it has some gaps related to class overlap and class imbalance. For this reason, the need of data transformation along with the selection of representative features is needed. 

\subsubsection{Data normalization.}
The high variability of the dataset values can be a huge problem that affects the quality of the ML models since it can introduce biasing \cite{fernandez2018learning}, which can lead to skewed learning algorithms. This may happen by the fact that these outliers may grab the attention disproportionately. To avoid this, a common solution is normalizing data. For this, normalization with the rule indicated in Expression \ref{eq:norm} is performed in this work.
\begin{equation}\label{eq:norm}
    \frac{X-X_{\min}}{X_{\max}-X_{\min}}
\end{equation}

In this case, normalization is applied to each particular feature, in order to give the same representation to each.

\subsubsection{Feature Selection.}
As it is proposed in \cite{zhang2018effective}, the selected features for this study have been \textit{sload, dload, dmeansz, smeansz, stcpb, dtcpb, sttl, djit} and \textit{trans-depth} and the dataset has been filtered by the feature \textit{service}, taking into account only the \textit{unknown, ftp} and \textit{dns} services. Before this process is done, the dataset has $2,540,044$ samples, and after the whole process, $2,045,019$ are left.

%\subsection{Dataset limitations}

\subsection{Evaluation Metrics}\label{EM}
For evaluating our solution and in order to compare the results with other state of the art technologies, several metrics are extracted in terms of performance, as well as disk usage and training time. The time is measured by capturing the needed time for the model to finish the training phase. 
This might be particularly interesting in a scenario in which the training is done in place, i.e. directly in the limited resource device. When talking about the NBcoded pipeline, the training time for the autoencoder is measured, as well as the training time for the NB. Then both times are summed up to measure the training time of the proposed architecture. In terms of disk usage, once the models are trained, they are saved as binary files in disk to measure the disk usage. Once again, to measure the disk usage of the NBcoded, the disk usage of the trained encoder is summed with the disk usage of a trained NB.

\begin{figure}[]
    \centering
    \includegraphics[scale=0.5]{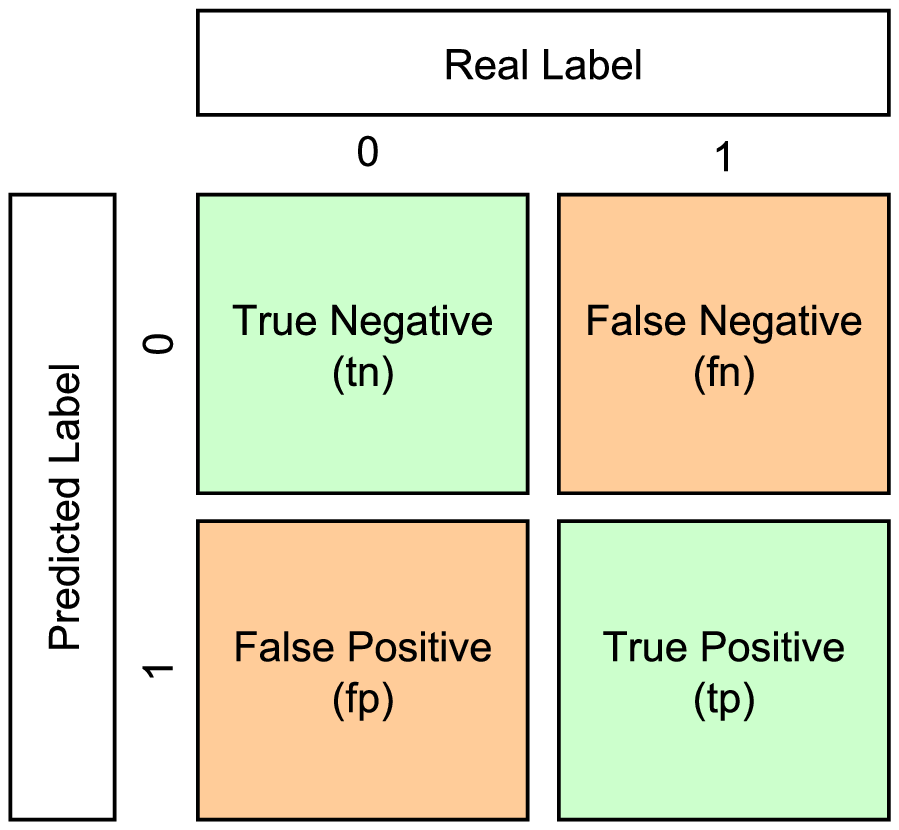}
    \caption{Feature matrix}
    \label{fig:featurematrix}
\end{figure}

The metrics related to the performance are the ones derived from the confusion matrix (Figure \ref{fig:featurematrix}) related to binary classification. 

These can be interpreted as it follows:

\begin{itemize}
    \item True Negative (tn): This is the case when the predicted value and the real value equals 0.
    \item False Negative (fn): This is the case when the predicted value is 0 but the real value equals 1.
    \item False Positive (fp): This one occurs when the predicted value is 1 but the real value equals 0.
    \item True Positive (tp): This one happens when both values equal 1. 
\end{itemize}
 From these values, some interesting metrics can be extracted, which represent well the performance of the models, that are
 \begin{itemize}
    \item Precision: This measures the relative success rate of the model referred to the total amount of real positives. It is calculated by
    $$
    \frac{tp}{tp+fn}.
    $$
    \item Recall: This measures the relative success rate of the model referred to the total amount of predicted positives. It is calculated by
    $$
    \frac{tp}{tp+fp}.
    $$
    \item Accuracy: This one measures the relative success rate of the model for the whole set. It can be computed by
    $$
    \frac{tp+tn}{tp+fp+tn+fn}.
    $$
    \item F1-score: The harmonic mean of the precision and recall. It is computed by
    $$
    2\cdot\frac{precision \cdot recall}{precision+recall}.
    $$
\end{itemize}

For the testing purposes, a stratified $10$-fold cross-validation is used, in splits of $80\%$ for the training sets and a $20\%$ for the tests sets which allows similarly balanced sets. Once the $10$ training and testing phases are completed, a mean and standard deviation of the results are calculated. 

\subsection{Machine learning models parameters and computer characteristics}\label{MLap}
The autoencoder is trained with a callback of 5 iterations is established for 100 epochs, in batches of 250.

For the other state-of-the-art methods, a 100 treed \cite{liu2015number} Random forest is chosen, with no limit of node expansion. Concerning the MLP, a two hidden layered model is chosen, where each of the layers is composed of 100 neurons. This one is trained with an early stop of 5 iterations within 100 epochs, in batches of 250.
The experiments where carried on a Windows 10 machine, with $8$GB of RAM and 4 CPUs of $3.40$GHz

\section{Experimental Study}\label{ExpSt}
In this section, the main concepts of the experiments carried in this study are exposed. Those are comprised by three experiments, where the aim of the first one is to create baseline values, the second one is to prove an optimization of these values and the third one aims to show a competitive performance of the proposed model in comparison with other state-of-the-art methods.

\subsubsection{Experiment 1.}\label{Exp1}
The aim of this first experiment is to compare different Gaussian NB, Complement NB, and Bernoulli NB models, in order to create baseline values and see the improvements of these in the next experiments. For this purpose, the three different NB classifiers are trained with the same training data and tested using the same test data, only in terms of \textit{precision}, \textit{recall}, \textit{accuracy} and \textit{F1}-score. As the experiment aims to establish baseline values for the three different NB classifiers, no metrics related to time and resource usage are taken into account.

\subsubsection{Experiment 2.}\label{Exp2}
The aim of this second experiment is to demonstrate the improvement of the previously tested algorithms when applying an encoding layer first. For this, the same $3$ NB models are studied again, in a stratified $10$-fold cross-validation, but in this case, $2$ training sets are subtracted, where each has $40$\% of the total dataset (and again a $20\%$ for the testing sets). In this split, an AE is trained with one of the training sets and then, encoder layers are subtracted from it. For the next step, the other training set is given to the encoder for the generation of the less dimensional dataset. Once the encoding is done, the $6$ featured dataset is used to feed the three NB models, in order to demonstrate an improvement of these, as well as to select the most competitive one. For the complement case, the dataset is translated to get positive values. Finally, the test set is given to the encoder, and the output of it is given to the NB models to classify samples into the attack or normal class.

\subsubsection{Experiment 3.}\label{Exp3}
Finally, in this third experiment, the model obtaining the best performance in the previous experiment is compared with state of the art technologies like RF and MLP, which are frequently used as a baseline \cite{wankhede2018attack}. As the aim of this experiment is to compare not only the general performance of these models, but the resource and time usage too, in this last experiment an evaluation not only measuring the \textit{precision}, \textit{recall}, \textit{accuracy} and \textit{F1}-score is carried, but also resource and time consumption are reported in terms of disk usage and training time.

\section{Results}\label{Results}
In this section, the results of the previously described experiments are presented.
\subsubsection{Experiment 1.}
These results will establish the starting point and the reference which with to compare when performing the experiments with the proposed tool. As it can be seen in Table \ref{tab:exp1}, the one getting better results is the Gaussian NB, which gets quite competitive results when this particular preprocess of the data is carried out, even without the application of an encoder layer.

\begin{table}[]
\centering
\begin{tabular}{lcccccccc}
\hline
\textbf{Models} & & \textbf{Precision} & & \textbf{Recall} & & \textbf{Accuracy} & & \textbf{F1} \\ \hline
\begin{tabular}[c]{@{}l@{}}GNB\end{tabular} & & $0.88\pm 0.01$ & & $0.98\pm 0$ & & $0.98\pm 0$ & & $92.8\%\pm 0\%$ \\ [2ex] 
\begin{tabular}[c]{@{}l@{}}CNB\end{tabular} & & $0.46\pm 0.03$ & & $0.92 \pm 0$ & & $0.84\pm 0.01$ & & $61.4\%\pm 2.7\%$ \\[2ex] 
\begin{tabular}[c]{@{}l@{}}BNB\end{tabular} & & $0.50\pm 0$ & & $0.88 \pm 0$ & & $0.87\pm 0$ & & $64.2\%\pm 0\%$ \\[2ex] 
\hline
\end{tabular}
\caption{Different NB results}
\label{tab:exp1}
\end{table}

\subsubsection{Experiment 2.}
In this experiment, the results of the NBcoded tool are shown, assuming Gaussian, Complement and Bernoulli distributions.  
\begin{figure}[]
    \centering
    \includegraphics[scale=0.5]{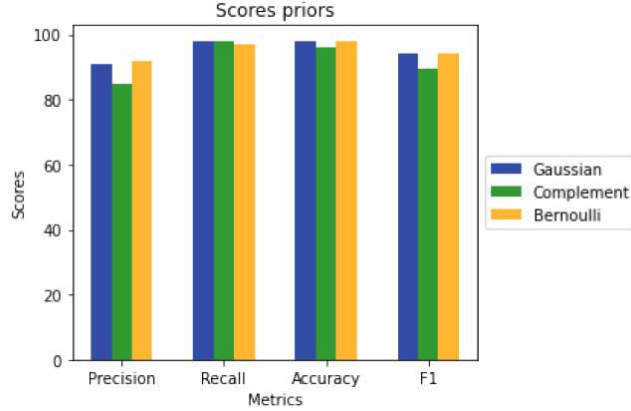}
    \caption{Results of different NB distribution combined with an encoder layer}
    \label{fig:exp2}
\end{figure}

As it can be seen in Fig \ref{fig:exp2}, the three models have had an improvement when classifying over the baseline values established in the first experiment. As it can be observed, the one getting better results has been the GNB, which has gotten scores of $0.91, 0.98, 0.98$ and $94.3\%$ in the \textit{precision, recall, accuracy} and \textit{F1-score} metrics respectively. In comparison with the first experiment, the Gaussian NB has gotten an improvement of $1.5\%$ in the F1-score, whereas the Complement NB and the Bernoulli NB, have gotten an improvement of $28.2\%$ and $30\%$ in the F1-score respectively, making the BNB the one with the most improvement, although the GNB is the one with the best performance. 
%, which has gotten scores of $0.91$, $0.98$, $0.98$ and $94.3\%$ in the \textit{precision}, \textit{recall}, \textit{accuracy} and \textit{F1}-score metrics respectively.

% \begin{table}[]
% \centering
% \begin{tabular}{lcccccccc}
% \hline
% \textbf{Models} & & \textbf{Precision} & & \textbf{Recall} & & \textbf{Accuracy} & & \textbf{F1} \\ \hline
% \begin{tabular}[c]{@{}l@{}}Gaussian NBcoded\end{tabular} & & $0.91\pm 0$ & & $0.98\pm 0$ & & $0.98\pm 0$ & & $94.3\%\pm 0\%$ \\ [2ex] 
% \begin{tabular}[c]{@{}l@{}}Complement NBcoded\end{tabular} & & $0.85\pm 0.18$ & & $0.98 \pm 0.0$1 & & $0.96\pm 0.08$ & & $89.6\%\pm 14.3\%$ \\[2ex] 
% \begin{tabular}[c]{@{}l@{}}Bernoulli NBcoded\end{tabular} & & $0.92\pm 0.01$ & & $0.97 \pm 0.01$ & & $0.98\pm 0$ & & $94.2\%\pm 0\%$ \\[2ex] 
% \hline
% \end{tabular}
% \caption{Results of different NB priors combined with an encoder layer}
% \label{tab:exp2}
% \end{table}

% As it can be seen in Table \ref{tab:exp2}, the three models have had an improvement when classifying. The Gaussian NB has gotten an improvement of 1.5\% in the F1-score, whereas the Complement NB and the Bernoulli NB, have gotten an improvement of 28\% and 30\% in the F1-score respectively, making the BNB the one with the most improvement, although the GNB is the one with the best performance. 

\subsubsection{Experiment 3}
For this last experiment, the comparison between the Gaussian NBcoded, the MLP and the RF is performed, in terms of the used 4 metrics, time and disk usage. 

\begin{table}[]
\centering
\begin{tabular}{lcccccccc}
\hline
\textbf{Models} & & \textbf{Precision} & & \textbf{Recall} & & \textbf{Accuracy} & & \textbf{F1} \\ \hline
\begin{tabular}[c]{@{}l@{}}Gaussian NBcoded\end{tabular} & & $0.91\pm 0$ & & $0.98\pm 0$ & & $0.98\pm 0$ & & $94.3\%\pm 0\%$ \\ [2ex] 
\begin{tabular}[c]{@{}l@{}}Multi Layer Perceptron \end{tabular} & & $0.92\pm 0.01$ & & $0.99\pm 0.01$ & & $0.99\pm 0$ & & $95.2\%\pm 0.01\%$ \\[2ex] 

\begin{tabular}[c]{@{}l@{}}Random Forest \end{tabular} & & $0.94\pm 0.01$ & & $0.99\pm 0.01$ & & $0.99\pm 0$ & & $96.5\%\pm 0\%$ \\[2ex]

\hline
\end{tabular}
\caption{Comparison of the Gaussian NBcoded, the MLP and the RF}
\label{tab:exp3}
\end{table}

As it is shown in Table \ref{tab:exp3}, the results obtained by the Gaussian NBcoded can compete in terms of precision, recall accuracy, and F1-score against the MLP and the RF. Moreover, in terms of training time and disk usage, the Gaussian NBcoded outperforms the two others, as it can be observed in Table \ref{tab:exp3_time}.

\begin{table}[]
\centering
\begin{tabular}{lcccc}
\hline
\textbf{Models} & & \textbf{Training time} & & \textbf{Disk usage} \\ \hline
\begin{tabular}[c]{@{}l@{}}Gaussian NBcoded\end{tabular} & & $169.15\pm 23.92$ & & $15.41\pm 0$ \\ [2ex] 
\begin{tabular}[c]{@{}l@{}}MLP \end{tabular} & & $282.89\pm 182.36$ & & $45.62\pm 0.16$  \\[2ex] 

\begin{tabular}[c]{@{}l@{}}RF \end{tabular} & & $217.97\pm 18.83$ & & $94,047.34 \pm 541.95 $ \\[2ex] 

\hline
\end{tabular}
\caption{Needed time in seconds for training, and disk usage in kilobytes.}
\label{tab:exp3_time}
\end{table}

\subsection{Discussion}
As it can be observed in the first experiment, the GNB gets quite good results detecting attacks when this particular selection of features is performed. However, the CNB and GNB do not fit well with the data. Even the recall and accuracy can be quite good in both cases, the remaining two measures are far from being competitive. In fact, due to the class imbalance found in the UNSW-NB15, this can be explained straightforwardly; the Bernoulli and Complement  NB models classify in a high rate attacks as normal behaviour. However, this scenario changes when the encoder comes into play. This might be due to firstly, the noise reduction in the data. Secondly, the encoder might model well the dependency between variables printing those in a less dimensional space. This may help the NB classifiers in their task, by the fact that the error given by the assumption of independence between variables might be reduced. This last statement could be reinforced due to the fact that using the encoder with the MLP and RF does not provide any advantage. In terms of training time, the Gaussian NBcoded seems to be the fastest one, even if this phase might be performed ``outside" of the limited system, i.e in a more suitable one, and finally transferring the trained model.

%When applying this first layer, the three classifiers improve their performance in all the measures getting much better results. Moreover, when comparing the best one to other well known ML algorithms, similar results are obtained in terms of precision, recall, accuracy, and F1, outperforming those in terms of training time and disk usage.

\section{Conclusions and Future Works}\label{Conc}
In this work, the idea was to create a novel attack classifier that can be used in limited resourced scenarios and without affecting the performance of small devices, as for example the ones used in IoT or Industry process. The classifier is tested against the UNSW-NB15 dataset, obtaining an F1-score of $94.3\%$, while maintaining a low $15.41$kb disk usage and a low $169.15$s of training time, which demonstrates the viability of the proposed classifier in attack classification tasks for low resourced devices. Future works can  compress the testing of the model in IoT devices, to ensure that the models do not interfere with their functionalities. Another interesting future approach might be to stack autoencoders to try to clean the data even more. This would add some training time, while keeping the disk usage, due to the fixed architecture of the encoder.  

\subsubsection*{Acknowledgements.} This work has been partially supported by the Spanish Centre for the Development of Industrial Technology (CDTI) under the project ÉGIDA (EXP 00122721 / CER-20191012) - RED DE EXCELENCIA EN TECNOLOGIAS DE SEGURIDAD Y PRIVACIDAD and by the Basque Country Government under the ELKARTEK program, project TRUSTIND (KK-2020/00054).


\bibliographystyle{splncs04}
\bibliography{NBcoded}

\end{document}